\documentclass[10pt,twocolumn,letterpaper]{article}

\usepackage{3dv}
\usepackage{times}
\usepackage{epsfig}
\usepackage{graphicx}
\usepackage{amsmath}
\usepackage{amssymb}
\usepackage{authblk}
\usepackage{array, makecell}
\usepackage{adjustbox} 
\usepackage{caption} 
\usepackage{booktabs} 
\usepackage{pifont}
\usepackage{subfig}
\newcommand{\OK}{\ding{51}}

\threedvfinalcopy 

\def\threedvPaperID{263} 

\ifthreedvfinal\fi

\begin{document}

\title{Introducing Pose Consistency and Warp-Alignment for Self-Supervised 6D Object Pose Estimation in Color Images}

\author{Juil Sock$^{1}$, Guillermo Garcia-Hernando$^{1,2}$, Anil Armagan$^{1}$, Tae-Kyun Kim$^{1,3}$}
\affil{$^1$Imperial College London \quad$^2$Niantic \quad$^3$KAIST\\

}

\maketitle


\begin{abstract}
Most successful approaches to estimate the 6D pose of an object typically train a neural network by supervising the learning with annotated poses in real world images. These annotations are generally expensive to obtain and a common workaround is to generate and train on synthetic scenes, with the drawback of limited generalisation when the model is deployed in the real world.
In this work, a two-stage 6D object pose estimator framework that can be applied on top of existing neural-network-based approaches and that does not require pose annotations on real images is proposed. The first self-supervised stage enforces the pose consistency between rendered predictions and real input images, narrowing the gap between the two domains.
The second stage fine-tunes the previously trained model by enforcing the photometric consistency between pairs of different object views, where one image is warped and aligned to match the view of the other and thus enabling their comparison. 
In the absence of both real image annotations and depth information, applying the proposed framework on top of two recent approaches results in state-of-the-art performance when compared to methods trained only on synthetic data, domain adaptation baselines and a concurrent self-supervised approach on LINEMOD, LINEMOD OCCLUSION and HomebrewedDB datasets.
\end{abstract}

\vspace{-0.5cm}
\section{Introduction}
\thispagestyle{empty}

6D object pose estimation~\cite{sahin2020review} is an emerging and important topic in the intersection of computer vision, robotics and augmented reality. State-of-the-art approaches have shown to perform reasonably well in real cluttered environments when only utilising colour images~\cite{brachmann2016uncertainty}. An obvious recent trend is the use of deep neural networks (DNNs), which have shown strong performance when annotated training data is available at scale~\cite{rad2017bb8,Park_2019_ICCV}.

To train DNN-based approaches, large quantities of training images can be synthesised where pose labels are automatically generated by rendering 3D object models in the desired poses. However, as we show in our experiments, naively training on synthetic images shows poor generalisation to the real domain~\cite{Zakharov_2019_ICCV,sundermeyer2018implicit}. Although visually similar, rendered images and real images are different in many ways due to multiple factors from compression effects to lens vignetting. Therefore DNNs often exploit real images with accurate 6D pose labels~\cite{tekinRealTimeSeamlessSingle2017,Park_2019_ICCV,rad2017bb8,li2019cdpn} in addition to the synthetic data. Unfortunately, it is cumbersome to obtain real data with high quality 6D pose annotations in large quantities. To overcome this limitation, supervised~\cite{rad2018feature} and unsupervised~\cite{bousmalis2017unsupervised,ganin2016domain} domain adaptation as well as self supervised methods~\cite{wang2020self6d,wan2019self} have been proposed to learn from both real and synthetic domain. However, domain adaptation either requires real pose labels(for supervised) or the performance is sub-optimal(for unsupervised). Furthermore, existing methods often require extra data other colour images to overcome the domain gap.

\begin{figure}[t]
\begin{center}
		\includegraphics[width=\linewidth]{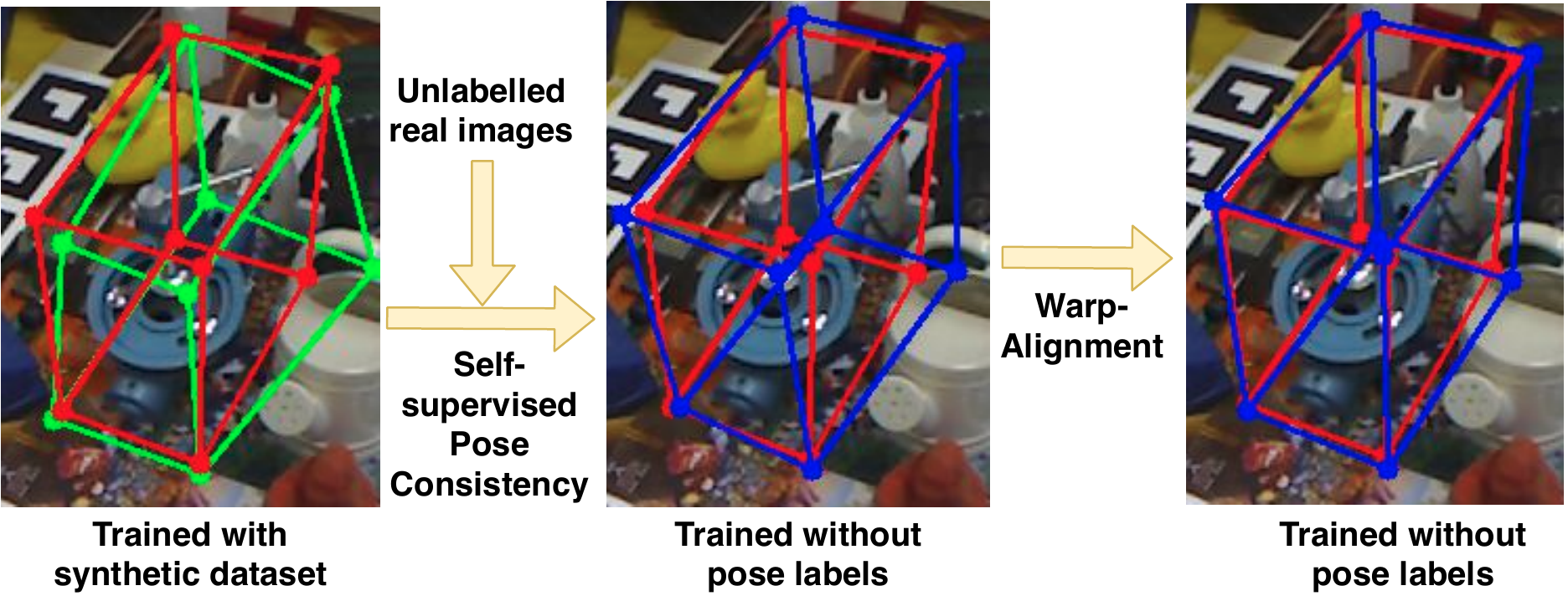}
    \caption{\textbf{ Illustration of the proposed framework.} We propose a self-supervised approach that leverages a pose consistency loss and warp-align stage to learn a robust 6D object pose estimator without using real pose labels. Red boxes show the ground-truth pose, and green and blue boxes represent estimated pose from a model trained with synthetic data only and the proposed self-supervision respectively.}
   
	\label{fig:qualitative}
\end{center}
\end{figure}

In this paper, we propose a 6D object pose estimation framework that can be learned without pose labels in the real domain. This framework consists of two main consecutive stages: a self-supervised stage that aims to bridge the synthetic-real gap, and a warp-alignment stage that adjusts the previous prediction by enforcing visual consistency between different object views.

The self-supervised stage uses the \textit{object pose consistency} as the supervisory signal, inspired by the cycle consistency approach introduced by Zhu et al.~\cite{zhu2017unpaired}. It guides the model towards learning to produce consistent pose estimations for both real and synthetic images. This is achieved by generating image-label pairs in the synthetic domain conditioned on the pose predictions in the real domain using a differentiable renderer during the training. 
The proposed approach works with RGB images only, as opposed to previous self-supervised methods that require extra modalities such as depth maps for fitting predictions to data~\cite{wang2020self6d}. This is desirable as capturing depth information requires special hardware which is not always available, particularly in outdoor applications. 

In the warp-alignment stage, we propose to fine-tune the previously trained model by exploiting the relation between the camera poses and the object poses. Inspired by approaches in monocular depth estimation literature~\cite{Ranjan_2019_CVPR,godard2019digging}, our observation is that the camera poses can be inferred from the predicted object poses to warp the input images into another view, to be perceptually compared and provide a supervisory signal to the pose estimator.

The proposed framework is flexible, and virtually any object pose estimator could benefit. We apply our framework on top of two different state-of-the-art methods, namely BB8~\cite{rad2017bb8} and Pix2Pose~\cite{Park_2019_ICCV}, and observed a boost in performance in the absence of pose annotation on real images. Moreover, we report state-of-the-art performance when compared to recent approaches that are trained on synthetic data, domain adaptation baselines and the aforementioned self-supervised method in the absence of both real pose annotations and depth information on LINEMOD~\cite{hinterstoisser2012model}, LINEMOD OCCLUSION~\cite{hinterstoisser2012model} and HomebrewedDB~\cite{Kaskman_2019_ICCV_Workshops} datasets.

\begin{figure*}[ht!]

\begin{center}
		\includegraphics[trim={2cm 0 0cm 0}, clip, width=1.\textwidth]{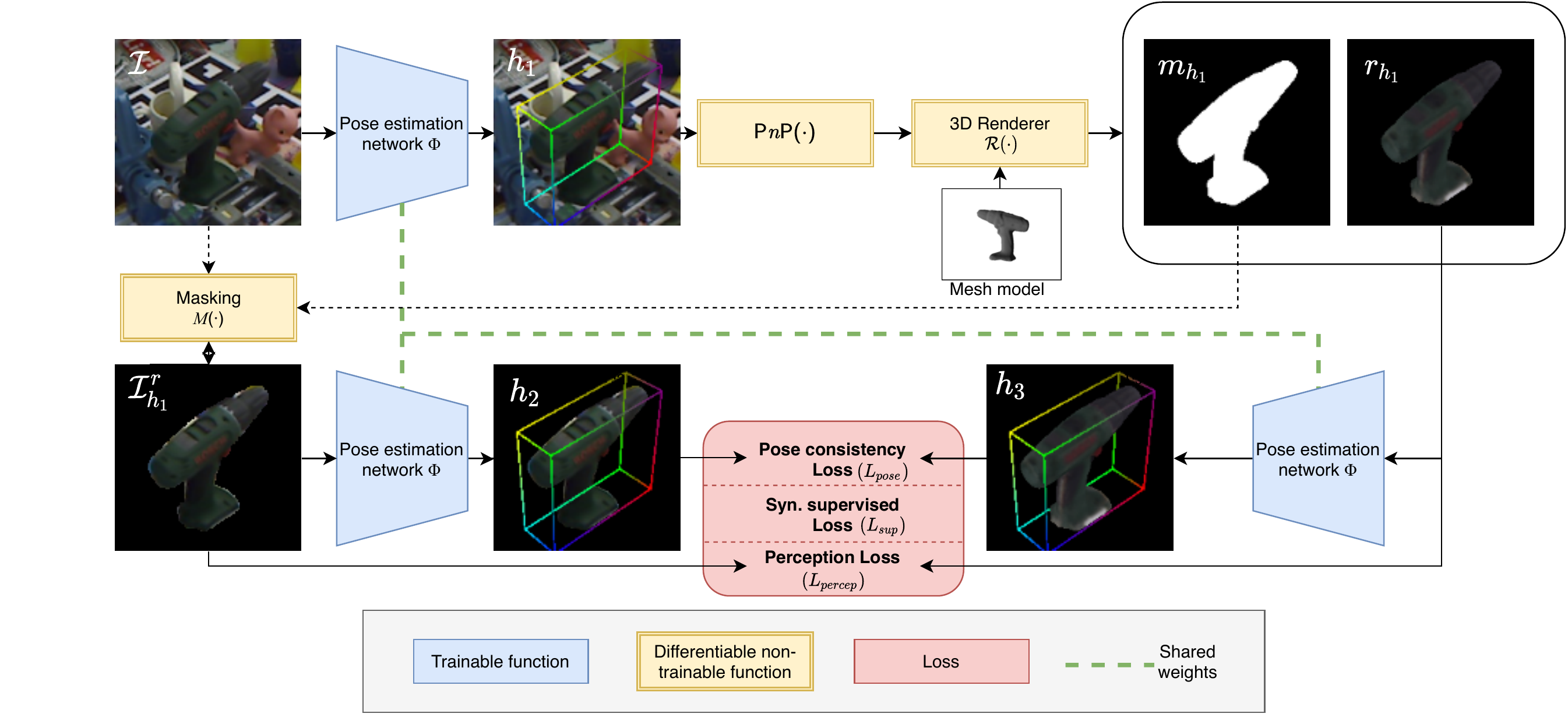}
    \caption{
    \textbf{Proposed self-supervised framework}.
    The figure shows the proposed self-supervised framework using pose consistency. We start training from a baseline pose estimator trained purely with synthetic RGB data. Given real unlabelled data, the pose is estimated and an image is rendered through a differentiable renderer based on the prediction. Object poses of input images augmented with the estimated silhouettes and rendered images are enforced to be consistent.
    }
	\label{fig:overview_1}
\end{center}
\end{figure*}

\section{Related Work}

\noindent\textbf{Datasets.} Annotating geometric labels such as 6D object pose is costly, often involving special hardware~\cite{yuan2017bighand2,h36m_pami}, and only captures limited views as the space of possible viewpoints is immense. Efforts have been made to create a real RGB dataset with high-quality labels such as Pix3D~\cite{pix3d}, TLESS~\cite{hodan2017tless}, YCB~\cite{xiang2017posecnn}, Bin-picking~\cite{doumanoglou2016recovering} and HomebrewedDB~\cite{Kaskman_2019_ICCV_Workshops}, yet have been limited due to the time and effort required.\\
\noindent\textbf{Real-synthetic domain gap.} Existing methods render synthetic images from a 3D model with perfect labels to train the network in the synthetic domain and deploy them in real-world ~\cite{sock18multi,Kehl_2017_ICCV}. However, training purely with synthetic images does not show good generalisation to real images~\cite{Zakharov_2019_ICCV}. Although visually similar, the rendered and real images are different in many ways due to factors such as compression effects or lens vignetting. To overcome this shortcoming, mainly two different approaches have been proposed for 6D object pose estimation: domain randomisation~\cite{zakharov2019deceptionnet,sundermeyer2018implicit} and domain adaptation~\cite{bousmalis2017unsupervised}. Domain randomisation hypothesises that enough variability in augmentation for simulated images will generalise to real images~\cite{tobin2017domain}.
However, the domain randomisation is limited to the type of parameters (e.g. brightness, contrast, etc.) being randomised, selected heuristically. The chosen parameters might not be relevant to the domain gap. Recently, Zakharov \etal~\cite{zakharov2019deceptionnet} proposed a method that learns the optimal weights for augmentation modules which helps achieve maximum domain confusion.
Domain adaptation is categorised into unsupervised~\cite{bousmalis2017unsupervised,ganin2016domain} and supervised~\cite{rad2018feature} learning. Supervised domain adaptation uses image-label pairs from both the source and target domain and directly learns the mapping between the two representations~\cite{georgakis2018matching,rad2018feature}. These methods show promising results but require labels in the target domain. Recently, unsupervised domain adaptation with Generative Adversarial Networks (GANs)~\cite{hoffman2017cycada,bousmalis2017unsupervised,murez2018image} has been proposed to generate target domain images without labels. Another example of an unsupervised method is GRL~\cite{ganin2016domain}, where domain-invariant features are generated to deceive domain classifiers. However, often the performance of these methods is suboptimal as the methods learn to match the distributions of the domains without considering the task at hand~\cite{Lee_2019_ICCV}. Also, such methods are prone to overfitting and degeneration of performance is observed if the samples are out of distribution~\cite{zakharov2019deceptionnet,sundermeyer2018implicit}.\\
\noindent\textbf{6D object pose estimation.} Recent 6D object pose estimators utilise only colour images as input and still achieve high accuracy when real data with label is used for training. The most popular approach is to estimate the correspondence between 2D images and 3D object models to obtain both 3D pose and translation via a Perspective-$n$-Point (P$n$P) algorithm~\cite{li2019cdpn,rad2017bb8,SPeng18_PVNet,Park_2019_ICCV,Wang_2019_CVPR}. One branch of such methods~\cite{rad2017bb8,tekinRealTimeSeamlessSingle2017,SPeng18_PVNet}, including BB8~\cite{rad2017bb8}, directly regresses pixel locations of 3D bounding box positions on 2D images where positions of 3D bounding boxes are predetermined. Alternatively, other methods~\cite{li2019cdpn,Park_2019_ICCV,Wang_2019_CVPR, sock2019active} including Pix2Pose, estimate normalised object coordinate values for every input image pixel to establish the 2D--3D correspondences. In this work, we chose one representative method from each branch as a baseline for our self-supervised and fine-tuning framework.\\
\noindent\textbf{Self-supervised pose estimation.} Recently, self-supervised methods have shown that a model can be trained to predict hand or object pose by constraining the prediction to be consistent with input data through fitting~\cite{mees2019self,palazzi2018end,wan2019self,wang2020self6d}. Although they do not require pose labels, they use an alternative form of ground truth as a weak supervisory signal. \cite{wang2020self6d} uses visual and depth alignment to learn 6D pose estimation without labels, however through experiment it has been shown that the accuracy largely relies on the geometric information from the depth map. Supervision from visual alignment does not provide reliable supervision as when the real image and the rendered image of the predicted pose are aligned, and the domain gap between them makes the supervision less meaningful. \cite{wan2019self} exploits both depth maps and multi-view data to achieve high performance. Self-supervised learning has also been used to learn meaningful feature representations by applying simple geometric augmentation to input images such as rotation~\cite{tran2019improved,chen2019self}. In our work, we propose to apply the silhouette of predicted pose to augment the input image, and this technique is shown to be crucial for bridging the synthetic--real domain gap in our experiments.\\
Another domain where self-supervision is widely used is monocular depth estimation from video~\cite{Ranjan_2019_CVPR,godard2019digging,Zou_2018_ECCV}. The idea is to estimate both the camera viewpoint and depth map for two consecutive frames, which enables the warping of the source frame to the target frame. A photometric loss between the warped frame and the target frame can serve as a proxy loss. Inspired from this idea, we propose an unsupervised learning framework which further enhances the pose estimator trained with our self-supervision. The core observation is that the relative viewpoint between two images can be calculated from the estimated poses. Also, rendering a depth map using the object model with an estimated pose via a differentiable renderer can provide geometric data needed without explicitly predicting pixel-wise depth values.

%

\section{Proposed Framework}
\subsection{Overview}
We propose a self-supervised method which results in a domain-invariant 6D pose estimator via {\em pose consistency} without pose labels on real RGB images. We assume that the primary differences between the two domains, real and synthetic, are low-level effects such as illumination and noise rather than high-level geometric variations~\cite{bousmalis2017unsupervised,kulkarni2019csm}. The proposed method adopts learning-by-synthesis in an end-to-end manner, via differentiable P$n$P and a differentiable renderer. The primary advantage of our proposed framework over the prior arts is that our method gets supervision from the discrepancy between the estimated poses on different domains, rather than getting a training signal from fitting. This eliminates the need for depth maps~\cite{wan2019self,wang2020self6d} which may not be available in common applications. In the second part of the section, we present a warp-align step that further enhances the accuracy of the pose estimator, again without requiring the real pose label.
\subsection{Self-supervised 6D Object Pose Learning}
\label{sec:method_self_sup}
Let $\{\mathcal{I}^g,y^g,\mathbf{h}^g\}$ denote a synthetic generated dataset and $\{\mathcal{I}^r\}$ a real one. $\mathcal{I}$ and $y$ represent images and object poses (3D rotation and translation), respectively. $\mathbf{h}$ is an intermediate representation, which is the output of a 6D object pose estimator with parameters $\theta$ and denoted $\Phi(\mathcal{I};\theta)$. The representation varies depending on the used baseline. For instance, $\mathbf{h}$ is the projected pixel locations of a 3D bounding box for the baseline BB8~\cite{rad2017bb8} and the normalised object coordinates and object mask for Pix2Pose~\cite{Park_2019_ICCV}. It can be computed as $\mathbf{h}^g=\pi(y^g,\mathcal{M},\mathbf{K})$, where $\pi(\cdot)$ is a function that maps the object pose to the intermediate representation for a given object model $\mathcal{M}$, and the camera intrinsic parameters $\mathbf{K}$. During training (see Fig.\ref{fig:overview_1}), the baseline pose estimator infers the intermediate representation from a real RGB image, $\mathbf{h}_1=\Phi(\mathcal{I}^r;\theta)$. $\mathbf{h}_1$ is used to yield $\mathcal{R}(\text{P}n\text{P}(\mathbf{h}_1,\mathbf{K}),\mathcal{M},\mathbf{K})=\{r_{\mathbf{h}_1},m_{h_1},y^g_{\mathbf{h}_1}\}$ via the P$n$P algorithm $\text{P}n\text{P}(\cdot)$ and the renderer $\mathcal{R}(\cdot)$. The functional outputs are the new synthetic image $r_{\mathbf{h}_1}$, its silhouette $m_{\mathbf{h}_1}$ and the estimated object pose $y^g_{\mathbf{h}_1}$. \\
A masking function $M(\mathcal{I}^r,r_{\mathbf{h}_1})=\mathcal{I}^r_{\mathbf{h}_1}$ outputs the masked real image using the silhouette. Finally the pose estimator predicts the intermediate representation for both the masked image $\mathcal{I}^r_{\mathbf{h}_1}$ and rendered synthetic image $r_{\mathbf{h}_1}$, that is $\Phi(\mathcal{I}^r_{\mathbf{h}_1};\theta)\rightarrow \mathbf{h}_2$ and $\Phi(r_{\mathbf{h}_1};\theta)\rightarrow \mathbf{h}_3$. As shown above, object poses are estimated multiple times across different domains, but the weights $\theta$ are shared. In the following, we present each component in detail:
\begin{figure*}[ht!]
\begin{center}
		\includegraphics[width=1.\textwidth]{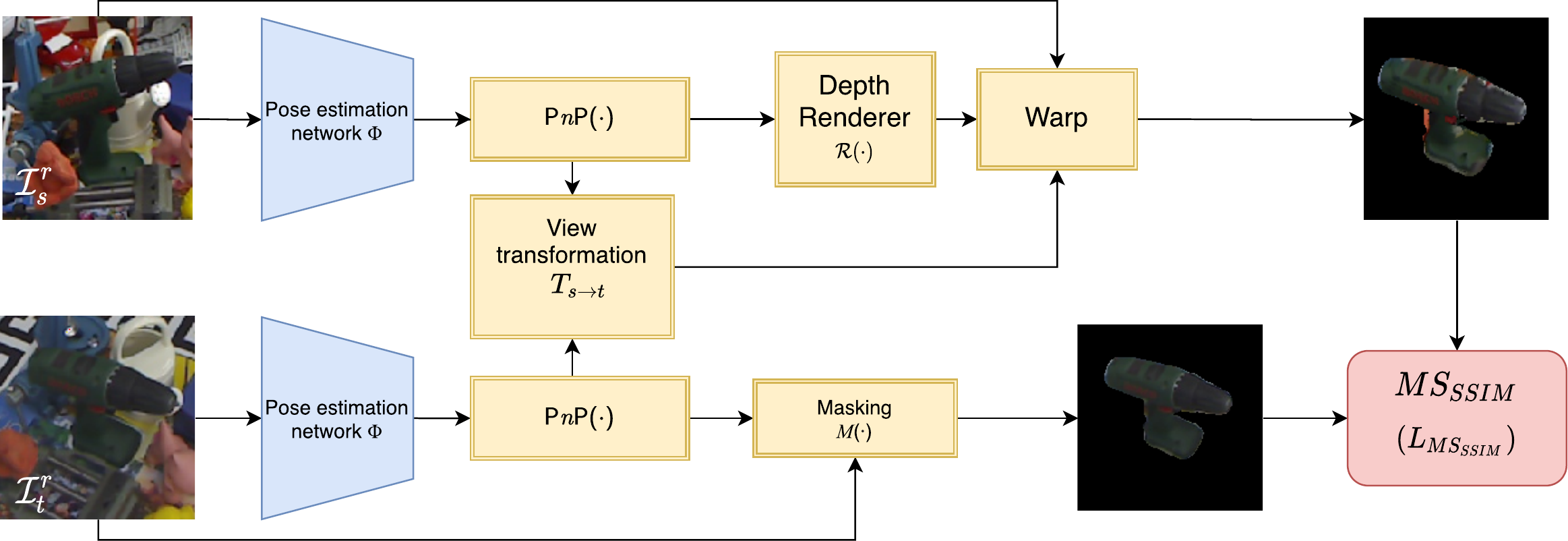}
    \caption{
    \textbf{Proposed warp-align stage}.
    Using a pair of unlabelled real images, predicted pose is used to estimate the relative camera pose. Together with depth rendered with a differentiable renderer based on the estimated pose, the source image is warped to the view point of the target image. Dissimilarity between the anchor and target image is used to update the pose estimator $\Phi$. The schematic follows the same colour coding as Fig.~\ref{fig:overview_1}.
    }
	\label{fig:overview_2}
	\vspace{-2em}
\end{center}
\end{figure*}

\noindent\textbf{Silhouette masking $M$.} Its function is to transform the real input image by removing the background with the silhouette rendered using the estimation $\mathbf{h}_1$: $M(\mathcal{I}^r,m_{\mathbf{h}_1}) = \mathcal{I}^r_{\mathbf{h}_1}$, where $M$ is the masking function to remove the background given a silhouette. The masked image is used to estimate the pose $\mathbf{h}_2$, which is encouraged to be consistent with $\mathbf{h}_3$. The intuition is that in order to achieve this goal, the estimated pose $\mathbf{h}_1$ needs to be correct, otherwise the foreground image would be removed resulting in the pose estimator not being able to estimate $\mathbf{h}_2$ consistently.

\noindent\textbf{Differentiable P${n}$P and Differentiable renderer $\mathcal{R}$.} In order to make the framework end-to-end trainable, we need a differentiable component to convert the output of the pose estimator to the 6D object pose.  We use the implementation from \cite{Chen_2020_CVPR} for differentiable P$n$P. We use a differentiable renderer~\cite{kato2018renderer,ravi2020pytorch3d}, denoted as $\mathcal{R}(\cdot)$, to generate synthetic images and silhouettes given the object model, camera parameters and object pose.

\noindent\textbf{Cross domain occlusion augmentation.} To make the pose estimator robust to occlusions, randomly sized patches with Gaussian noise are placed on both the rendered image $r_{\mathbf{h}_1}$ and target domain image $\mathcal{I}^r$ in each iteration during training. The part of the silhouette $m_{\mathbf{h}_1}$ occluded by the noise patch is excluded to avoid the masked real image $\mathcal{I}^r_{\mathbf{h}_1}$ containing the noise patch.

\noindent\textbf{Pose consistency loss.} The core idea is to encourage the pose estimator to estimate consistent poses for both synthetic and real domains. We define the pose consistency loss by penalising the discrepancy between the estimations from different domains. $L_{pose}$ enforces consistency of the transformed vertices of the object model based on the estimated poses, and measure the alignment in 3D space:
\begin{equation}
    L_{pose} = \|\text{P}n\text{P}(\mathbf{h}_2,\mathbf{K}){\mathcal{M}_v}-\text{P}n\text{P}(\mathbf{h}_3,\mathbf{K})){\mathcal{M}_v}\|_{2},
\end{equation}
where ${\mathcal{M}_v}$ are the positions of the vertices. $L_{pose}$ is agnostic to the baseline pose estimator $\Phi$ used, since the output of the pose estimator is converted to homogeneous transformation matrix through differentiable P$n$P.

\noindent\textbf{Synthetic domain supervised loss.} The synthetic domain supervised loss exploits the ground truth pose label that is available in the synthetic domain:
\begin{equation}
\begin{split}
    L_{sup} = \mathbb{E}[\|\Phi(r_{\mathbf{h}_1};\theta)-\pi(y^{g}_{\mathbf{h}_1},\mathcal{M},\mathbf{K})\|_{huber}]\\
    +\mathbb{E}[\|\Phi(\mathcal{I}^g;\theta)-\mathbf{h}^{g}\|_{huber}], 
    \end{split}
\end{equation}
where $r_{\mathbf{h}_1}$ and $\mathcal{I}^g$ are the rendered image via the proposed pipeline and the synthetic images in the training set and $huber$ denotes the Huber loss. %
The synthetic domain supervised loss helps to impose constraints on the real data self-supervision, as learning by the pose consistency alone can result in a singularity issue. For instance, the pose estimator can learn always to predict a trivial pose, e.g. frontal pose, regardless of the input data, in which the pose consistency loss is minimised.
It also helps to learn the prior geometric constraint, e.g.\ the output of the pose estimator always needs to form a valid bounding box in case the $\Phi$ is BB8. In general, this loss narrows the gap between synthetic supervision and real self-supervision, preventing the model parameter learning from diverging. 

\noindent\textbf{Overall training objective.} In addition to the expected loss over the consistency and the synthetic domain supervised loss, we also use an additional perceptual distance~\cite{zhang2018unreasonable} to encourage the rendered image $r_{\mathbf{h}_1}$ and the masked real image $\mathcal{I}^r_{\mathbf{h}_1}$ to be perceptually similar. We compute the $L_2$ distance in the feature space of pre-trained VGG~\cite{simonyan2014very} network between the two images where feature stack from multiple layers are normalised in
the channel dimension and denote this loss $L_{percep}$. The overall training objective is given as:
\begin{equation}
    \begin{split}
        L_{total} = \lambda_{pose}L_{pose}+\lambda_{sup}L_{sup}
        +\lambda_{percep}L_{percep},
    \end{split}
\label{eqn:total_loss_da}
\vspace{-2em}
\end{equation}
where $\lambda_i$ are constant weights to control the interaction of the losses. For estimator which separately estimate object mask (e.g.\ Pix2Pose), $L_1$ loss is used for both domains with the ground-truth mask. With the above objective, the proposed framework allows us to learn the pose estimator $\Phi$ without the pose labels in the target domain (i.e.\ real RGB images). At the test time, only $\Phi$ and P$n$P are used which is light-weight and fast.

\subsection{Warp-Align Fine-Tuning Stage}
Different to the self-supervised learning that learns to estimate poses by leveraging consistency in high-level features such as pose, the goal of the warp-alignment fine-tuning stage is to leverage consistency in lower-level photometric features.
Similar to the self-supervised framework, this step can be used with various off-the-shelf pose estimators without using any pose labels due to its unsupervised nature.
In Fig.~\ref{fig:overview_2} we show a schematic of the warp-alignment framework. A pair of unlabelled real input images, source image ($\mathcal{I}^r_{s}$) and target image ($\mathcal{I}^r_{t}$), are compared and the model learns to minimise the difference between the two images. However, since the two images are taken from different viewpoints, they cannot be compared directly. The core idea is to transform (warp) $\mathcal{I}^r_{s}$ to have the same viewpoint as $\mathcal{I}^r_{t}$, at which the images can be compared directly. 
In order to achieve the correct transformation and projection to 2D image, two ingredients are required: relative camera view transformation between the two images and depth value for the pixels of the image to be transformed, since the source image needs to be represented in 3D in order to apply 3D camera view transformation.\\
In 6D object pose estimation problems, the object pose is defined in the camera coordinate space. Hence the camera pose can be defined in the object coordinate space through a simple matrix inversion if the pose is represented as a homogeneous transformation matrix. Since the camera pose for both $\mathcal{I}^r_{s}$ and $\mathcal{I}^r_{t}$ can now be defined in the common object coordinate space, the relative view transformation is obtained as follows:
\begin{equation}
    T_{{s}\rightarrow{t}} = \text{P}n\text{P}(\Phi(\mathcal{I}^r_{t};\theta))(\text{P}n\text{P}(\Phi(\mathcal{I}^r_{s};\theta)))^{-1}.
\label{eqn:cam_pose}
\end{equation}

The depth values of the foreground pixels are required to warp the image to a different viewpoint and that can be obtained by rendering the 3D object model with the pose predicted by $\Phi$. The masking function $M$ is used to remove the background pixels.
3D points of the foreground pixels are then calculated using the rendered depth map and these points are later transformed by $T_{{s}\rightarrow{t}}$ to be projected onto an image plane.\\
We use the Multi-Scale Structural Similarity Index ($\mathsf{MS_{SSIM}}$)~\cite{MSSSIM} to calculate the similarity between the warped source image $\tilde{\mathcal{I}}^r_{s}$ and the target image $\mathcal{I}^r_{t}$, defined as follows:
\begin{equation}
     L_{\mathsf{MS_{SSIM}}} = 1 - \mathsf{MS_{SSIM}}(\tilde{\mathcal{I}}^r_{s}, M(\mathcal{I}^r_{t}, m_{\mathbf{h}_{t}})).
\label{eqn:cam_pose}
\end{equation}

\begin{table*}[t]
\centering
	\caption{ \textbf{Quantitative results on the LINEMOD dataset with ADD metric}. Results obtained with the BB8 baseline (Top). Results with the Pix2Pose baseline (Bottom).}
	\resizebox{\textwidth}{!}{
	\begin{tabular}{c|c|c|c|c|c|c|c|c|c|c|c|c|c||c}
	\toprule
	&Ape&Bvise&Cam&Can&Cat&Drill&Duck&Eggbox&Glue&Holep&Iron&Lamp&Phone&Mean\\
	\toprule
	BB8 Synth only(LB)&0.46&0.75&0.00&0.00&0.00&0.00&0.09&0.00&0.38&0.00&0.10&0.00&0.00&0.14\\
	\midrule
	BB8+GRL~\cite{ganin2016domain}&0.74&3.00&0.56&0.96&4.52&1.26&5.98&29.39&8.65&2.64&12.10&0.93&2.95&5.67\\
	BB8+PixelDA~\cite{bousmalis2017unsupervised}&1.40&0.66&1.02&0.10&0.66&0.19&0.18&0.00&11.85&0.91&3.60&1.39&4.00&1.98\\
	BB8+Ours&27.58&63.07&43.37&51.82&47.74&50.05&34.54&83.89&70.08&22.82&31.60&63.33&38.80&48.36\\
	\midrule
	BB8 with Real Pose Labels(UB)&24.79&63.07&46.43&61.61&46.39&55.86&27.11&88.62&84.29&34.27&76.90&79.16&55.48&57.23\\
	\bottomrule
	\multicolumn{15}{c}{\vspace{0.5cm}}\\

	\midrule
	&Ape&Bvise&Cam&Can&Cat&Drill&Duck&Eggbox&Glue&Holep&Iron&Lamp&Phone&Mean\\
	\toprule
	Pix2Pose Synth only(LB)&36.73&59.01&27.82&28.71&39.48&22.29&31.89&71.11&29.86&24.62&51.94&38.73&25.73&37.53\\
	\midrule
	Pix2Pose+Ours&39.68&72.19&56.54&53.88&50.05&31.54&36.49&83.86&53.24&33.84&76.63&73.35&52.13&54.88\\
	Pix2Pose+Ours+Alignment&37.62&78.67&65.45&65.55&52.59&48.86&35.09&89.26&64.54&41.55&80.90&70.74&56.47&60.56\\
	\midrule
	Pix2Pose with Real Pose Labels(UB)&54.61&90.21&72.75&90.66&76.47&78.91&54.41&99.53&89.28&75.76&97.04&95.59&79.56&81.14\\
	\bottomrule
	\end{tabular}
	}
	\label{tab:BB8_pix2pose_linemod}
\end{table*}

\section{Experiments}
We evaluate the efficacy of our framework through experiments on three datasets, and an ablation study is presented to investigate how different components contribute to the performance. We chose two baselines for $\Phi(\cdot)$, namely BB8~\cite{rad2017bb8} and Pix2Pose~\cite{Park_2019_ICCV}. These methods are essentially different, which makes them good candidates to demonstrate the generalisability of the proposed framework. BB8 is a holistic pose estimator which regresses pixel locations of a projected 3D bounding box. Pix2Pose uses object coordinates as an intermediate representation and learns to estimate foreground object pixels (mask), object coordinates and confidence scores. In other words, BB8 estimates the pixel location given 3D points of bounding box whereas Pix2Pose estimates a 3D point given a pixel location. Note that we use our own implementations for the baselines and both baselines achieve a better performance than the ones reported in the original works. Training the baseline purely on synthetic images serves as a \textbf{Lower Bound (LB)}, since the self-supervised training is initialised with the model pretrained with synthetic images. The baseline methods trained with both real and synthetic pose labels serves as an \textbf{Upper Bound (UB)}. 

We evaluate our approach on the commonly used benchmarks LINEMOD \cite{hinterstoisser2012model} and LINEMOD OCCLUSION~\cite{hinterstoisser2012model}, containing 13 and 8 objects respectively, as well as 3 objects from the HomebrewedDB dataset following Wang et al.~\cite{wang2020self6d}. All reported results are the percentage of correctly predicted poses using the AD\{D$\mid$I\}~\cite{hinterstoisser2012model} metric. We set a threshold of 10\% of the object's diameter following the BOP (Benchmark for 6D Object Pose Estimation) standard.

\subsection{Implementation Details}
For self-supervised learning, a batch size of 16 is used with the ADAM optimiser~\cite{kingma2014adam} for both baselines. The initial learning rate is 1e--5 for the first 15 epochs and reduced to 1e--6 for the next 10 epochs. 
We kindly refer to the supplementary material for details on parameters for each baseline used, as well as the parameters for the warp-alignment framework. For the LINEMOD dataset, following the standard practice in the community, 15\% of the images are used for training but \textit{without} pose labels. The selection of training images follows the strategy in~\cite{brachmann2016uncertainty}. To be robust against inaccurate detection, we randomly perturb the scale and center position of the bounding box during training. We initialise the network by training with synthetic data only where the early layers are frozen to avoid overfitting to the synthetic dataset. 
All parameters are updated after initialisation. For all testing, Faster-RCNN~\cite{ren2015faster} is used to detect objects and provides accurate bounding boxes with an accuracy of 99.24\% and 97.29\% with 50\% and 75\% IOU respectively, for all objects in the LINEMOD dataset. \\
Warp-align fine-tuning starts training from the pose estimator trained with self-supervised framework. Source images and target images are sampled such that the estimated 3D pose difference between the two is less than $60^{\circ}$, as warping into a very different viewpoint degrades the quality of the warped image.

\subsection{Results}
\noindent\textbf{Ablation Study}. Table~\ref{tab:da_ablation} shows the influence of different components for the self-supervised object pose estimation on camera object from the LINEMOD dataset with BB8 as the baseline method. It can be seen that silhouette masking is a critical component of the framework. Silhouette masking imposes a strong constraint as the pose consistency loss reduces if the first estimation $\mathbf{h}_1$ is closer to the correct pose, thereby not masking out a portion of the foreground. This intuition is further reinforced in our experiments. Table~\ref{tab:BB8_pix2pose_linemod} shows that the objects with more distinctive silhouettes for different poses (e.g. Benchvise) obtain higher accuracies than those that are not (e.g. Duck). The silhouette of a spherical object is invariant to the rotation, which means a wrong 3D pose estimation may still produce the correct silhouette. Perception loss makes training more stable, and occlusion augmentation further boosts the performance since it enforces the estimator to learn more robust features.

\begin{figure*}[t]
\begin{center}
		\includegraphics[width=1\linewidth]{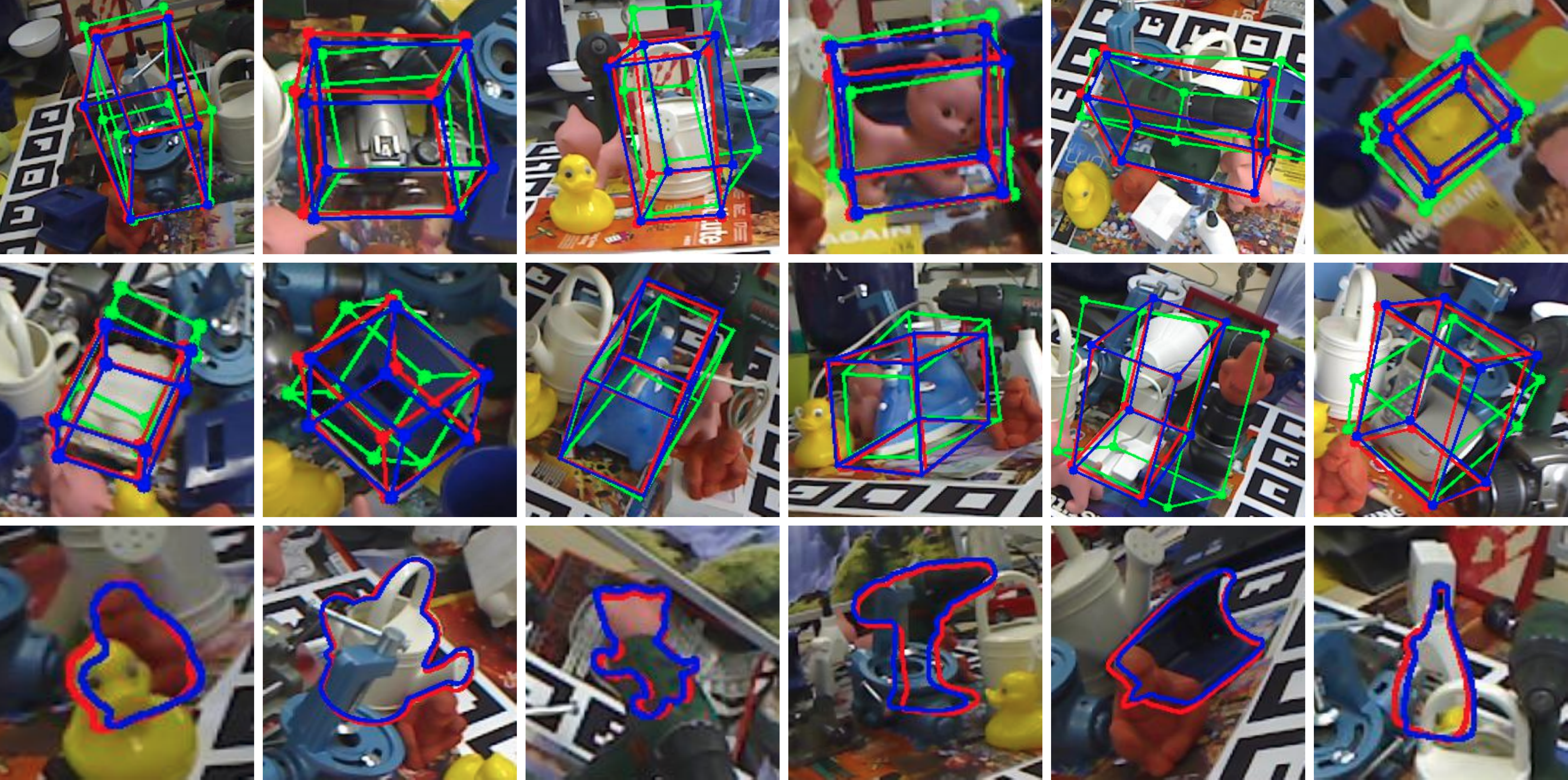}
    \caption{\textbf{ Qualitative results on LINEMOD and LINEMOD OCCLUSION.} First two rows show the visualisations with projected 3D bounding boxes for LB, Ours and ground truth on LINEMOD dataset. Red, blue and green boxes correspond to ground truth, self-supervised and LB respectively. 
    The last row visualises the estimated poses for LINEMOD OCCLUSION dataset. The blue and red silhouette corresponds to our prediction and ground truth.}
    \vspace{-1em}
	\label{fig:qualitative}
\end{center}
\end{figure*}

\begin{table*}[t]
    \centering
	\caption{\textbf{Comparison with state of the art.} Quantitative results for LINEMOD dataset by using Pix2Pose~\cite{Park_2019_ICCV} as our baseline pose estimator.}
	\resizebox{\textwidth}{!}{
	\begin{tabular}{c|cccccc|ccc}
			\toprule
			 Train data&\multicolumn{6}{c|}{\makecell{w/o Real Pose Labels}}& \multicolumn{3}{c}{\makecell{with Real Pose Labels}}\\
			 \midrule
			 Object&SSD~\cite{Kehl_2017_ICCV}&AAE~\cite{sundermeyer2018implicit}&MHP~\cite{manhardt2019explaining}&DPOD~\cite{Zakharov_2019_ICCV}&Self6D~\cite{wang2020self6d}&Ours&Tekin~\cite{tekinRealTimeSeamlessSingle2017}&DPOD~\cite{Zakharov_2019_ICCV}&CDPN~\cite{li2019cdpn}\\
			 \midrule
			 Ape&0.0&4.2&11.9&35.1&\textbf{38.9}&37.6&21.6&53.3&\textbf{64.4}\\
			 Bvise&0.2&22.9&66.2&59.4&75.2&\textbf{78.6}&81.8&95.2&\textbf{97.8}\\
			 Cam&0.4&32.9&22.4&15.5&36.9&\textbf{65.5}&36.6&90.0&\textbf{91.7}\\
			 Can&1.4&37.0&59.8&48.8&65.6&\textbf{65.6}&68.8&94.1&\textbf{95.9}\\
			 Cat&0.5&18.7&26.9&28.1&\textbf{57.9}&52.5&41.8&60.4&\textbf{83.8}\\
			 Drill&2.6&24.8&44.6&59.3&\textbf{67.0}&48.8&63.5&\textbf{97.4}&96.2\\
			 Duck&0.0&5.9&8.3&25.6&19.6&\textbf{35.1}&27.2&66.0&\textbf{66.8}\\
			 Eggbox&8.9&81.0&55.7&51.2&\textbf{99.0}&89.2&69.6&99.6&\textbf{99.7}\\
			 Glue&0.0&46.2&54.6&34.6&\textbf{94.1}&64.5&80.0&93.8&\textbf{99.6}\\
			 HoleP&0.3&18.2&15.5&17.7&16.2&\textbf{41.5}&42.6&64.9&\textbf{85.8}\\
			 Iron&8.9&35.1&60.8&84.7&77.9&\textbf{80.9}&75.0&\textbf{99.8}&97.9\\
			 Lamp&8.2&61.2&-&45.0&68.2&\textbf{70.7}&71.1&88.1&\textbf{97.9}\\
			 Phone&0.18&36.3&34.4&20.9&58.9&\textbf{60.5}&47.7&71.4&\textbf{90.8}\\
			 \midrule
			 \midrule
			 Mean&2.4&32.6&38.8&40.5&59.0&\textbf{60.6}&56.0&82.6&\textbf{89.9}\\
			 \bottomrule
			
		\end{tabular}}
		\label{tab:pix2pose_linemod_sota}
		\vspace{-1em}
\end{table*}

\begin{table}[t]
    \centering
	\caption{\textbf{Quantitative results on LINEMOD Occlusion.} All methods in the list do not use real pose labels during training. Results except ours are from~\cite{wang2020self6d}.}
	\resizebox{\linewidth}{!}{
	\begin{tabular}{c|ccc|c}
	\toprule
	Train data&\multicolumn{3}{c|}{\makecell{RGB image only}}& \multicolumn{1}{c}{\makecell{RGB+depth}}\\
	\midrule
	object&DPOD~\cite{Zakharov_2019_ICCV}&CDPN~\cite{li2019cdpn}&Ours&Self6D~\cite{wang2020self6d}\\
	\midrule
	Ape&2.3&\textbf{20.0}&12.0&13.7\\
	Can&4.0&15.1&\textbf{27.5}&43.2\\
	Cat&1.2&\textbf{16.4}&12.0&18.7\\
	Duck&10.5&5.0&\textbf{20.5}&32.5\\
	Drill&7.2&22.2&\textbf{23.0}&14.4\\
	Eggbox&4.4&\textbf{36.1}&25.1&57.8\\
	Glue&12.9&\textbf{27.9}&27.0&54.3\\
	HoleP&7.5&24.0&\textbf{35.0}&22.0\\
	\midrule
	Mean&6.3&20.8&\textbf{22.8}&32.1
	
	\end{tabular}
	}
	\label{tab:pix2pose_linemod_occ}
\end{table}

\noindent\textbf{Comparison with LB and UB}.
Table~\ref{tab:BB8_pix2pose_linemod} (Top) indicates that the LB performance for the BB8 baseline is very low with the ADD metric on LINEMOD. The middle rows show how unsupervised domain adaptation methods improve from the LB. Although PixelDA~\cite{bousmalis2017unsupervised} generates perceptually realistic images, it does not contribute to learning domain invariant features, as reported in~\cite{rad2018feature}. This can be explained by the fact that the number of real images allowed in the standard LINEMOD protocol (15\%) may not be enough to learn the distribution which results in overfitting on the training data. Using GRL~\cite{ganin2016domain} without real labels slightly improves the generalisability from synthetic training, as the only supervision is from the domain label which disregards the task. Our method significantly outperforms both methods and almost reaches the performance of the UB with real labels.\\
Table~\ref{tab:BB8_pix2pose_linemod} (Bottom) shows the performance on LINEMOD with Pix2Pose baseline. The performance with synthetic only training is much higher than BB8, and it can be explained by the fact that Pix2Pose regresses the object coordinates for every pixel and utilises a RANSAC framework which makes it more robust against false estimation, whereas BB8 regresses the location of eight corner points. Our proposed self-supervised learning framework improves significantly from the LB similar to the BB8 baseline experiment. Our warp-align stage boosts the performance further from the self-supervised framework except for some objects such as Ape, Duck or Lamp. Our alignment framework compares the texture between two images, but since those objects are almost monochromatic, alignment may not be able to provide meaningful supervision. Fig.~\ref{fig:qualitative} shows the qualitative results with LB, self-supervised and ground truth. It shows that the alignment with ground truth is improved after self-supervision.
\\
\textbf{Comparison with state-of-the-art}.
In Table.~\ref{tab:pix2pose_linemod_sota}, we compare our proposed work with other state-of-the-art methods. We distinguished the methods based on whether the methods are supervised with real pose labels or not. Our method significantly outperforms SSD~\cite{Kehl_2017_ICCV}, AAE~\cite{sundermeyer2018implicit}, MHP~\cite{manhardt2019explaining}, and DPOD~\cite{Zakharov_2019_ICCV} which all uses domain randomisation to learn from synthetic data. Notably, our method is on par with Self6D~\cite{wang2020self6d} which uses both visual and geometric alignment when utilising a depth map during training.\\
Table~\ref{tab:pix2pose_linemod_occ} reports the performance on the LINEMOD OCCLUSION~\cite{hinterstoisser2012model} dataset and all methods in the table do not use real pose labels during training. The results in the table follow the BOP~\cite{hodan2018bop} standard where 200 images from the original occlusion dataset are selected for the evaluation. Our method outperforms DPOD~\cite{Zakharov_2019_ICCV} and CDPN~\cite{li2019cdpn} which use RGB images during training, but performs lower than Self6D~\cite{wang2020self6d} when tested on images with strong occlusions, since using extra modalities such as depth in Self6D can help to learn more robust features in the presence of occlusions.\\
Table~\ref{tab:pix2pose_homebrewed} shows the results for the HomebrewedDB~\cite{Kaskman_2019_ICCV_Workshops} dataset. HomebrewedDB is a suitable dataset to demonstrate whether the trained pose estimator is not overfitted on the data used for training and generalisable, as 3 objects from LINEMOD dataset are captured in different backgrounds and lighting conditions. The object estimator trained with 15\% of unlabelled data from LINEMOD was tested on this dataset without retraining. The results show our self-supervised framework learns robust features that can generalise beyond the images in the training dataset. Our method outperforms DPOD~\cite{Zakharov_2019_ICCV} and SSD~\cite{Kehl_2017_ICCV} by 59\% and  20\% respectively.

\begin{table}[t]
    \centering
	\caption{\textbf{Quantitative results on HomebrewedDB~\cite{Kaskman_2019_ICCV_Workshops}}. All methods in the list do not use real pose labels during training. Results except ours are from~\cite{wang2020self6d}.}
	\resizebox{\linewidth}{!}{
	\begin{tabular}{c|ccc|c}
	\toprule
	Train data&\multicolumn{3}{c|}{\makecell{RGB image only}}& \multicolumn{1}{c}{\makecell{RGB+depth}}\\
	\midrule
	object&DPOD~\cite{Zakharov_2019_ICCV}&SSD~\cite{Kehl_2017_ICCV}+Ref&Ours&Self6D~\cite{wang2020self6d}\\
	\midrule
	Bvise&52.9&\textbf{82.0}&57.3&72.1\\
	Drill&37.8&22.9&\textbf{46.6}&65.1\\
	Phone&7.3&24.9&\textbf{41.5}&41.8\\
	\midrule
	Mean&32.7&43.3&\textbf{52.0}&59.7
	\end{tabular}
	}
	\label{tab:pix2pose_homebrewed}
\end{table}

\begin{table}[h]
    \centering
	\scriptsize
	\caption{\textbf{Ablation} study on how different terms in the loss and augmentations impact the performance on the camera object for the LINEMOD dataset.}
	\resizebox{\columnwidth}{!}{
	\begin{tabular}{cccc|c}
	    \toprule
	    Synthetic Supervision&Silhouette masking&Perception Loss&Occlusion& ADD \\
	    \midrule
	    \OK& & & &0.00 \\
	    \OK&\OK & & &29.66 \\
	    \OK&\OK &\OK & &35.31 \\
	    \OK&\OK &\OK &\OK &39.20 \\
	\end{tabular}}
	\label{tab:da_ablation}
\end{table}

\section{Conclusion}
We proposed a framework that consists of two main components: a self-supervised training approach and a warp-align fine-tuning stage. Our experimental results show that our framework can learn a robust 6D pose estimator without the use of real pose labels. Furthermore, we have also experimentally shown that the framework is flexible to use off-the-shelf object pose estimators as a baseline method by choosing two popular but distinctive approaches. Our evaluations on multiple datasets show that the proposed framework significantly improves on models trained with synthetic data only, and that it can generalise to different datasets without retraining. Finally, our method shows state-of-the-art performance in the absence of real data and depth information surpassing current self-supervised approaches.

{\small
\bibliographystyle{ieee}
\bibliography{egbib}
}

\clearpage

\section*{Supplementary material}
\setcounter{section}{0}

Pytorch~\cite{paszke2017automatic} is used to train and test the proposed method on a machine with Intel i5 and GTX1080TI. The objec models($\mathcal{M}$) used with the differentiable renderer($\mathcal{R}$)~\cite{kato2018renderer,ravi2020pytorch3d} were downsampled to reduce the training time if the number of faces are too large.
\paragraph{Self-supervision} Adam optimiser with The initial learning rate of 1e-5 was used for the first 15 epochs and  1e-6 for the next 10 epochs. Batch size of 16 was used. We followed the standard practice to generate training dataset~\cite{rad2017bb8,SPeng18_PVNet}. More specifically, we copy-and-pasted the object image onto random background images. During training, images were cropped 1.3 times the size of shorter side of bounding box. The mean position of bounding box was perturbed by Gaussian distribution of mean value of 0 pixel and standard deviation of 3 pixel and randomly rescaled. Input images were augmented by: adding Gaussian noises, Contrast normalisation and Gaussian blur.\\
\paragraph{Warp-alignment} Adam optimiser with learning rate of 0.0001 was used where learning rate were droped by 0.1 every 25 epoches. For every minibatch, 25 images of real images were sampled from the training dataset which produced 30300 possible pair combinations between the sampled images. Among them, 25 pairs with the minimum estimated 3D pose difference were selected to form a minibatch of batchsize 25.

\end{document}